\documentclass[11pt]{article}

\usepackage[T1]{fontenc}
\usepackage{lmodern}
\usepackage[english]{babel}
\usepackage{microtype}
\usepackage[a4paper,margin=0.85in]{geometry}
\usepackage{graphicx}
\usepackage{booktabs}
\usepackage{tabularx}
\usepackage{multirow}
\usepackage{array}
\usepackage{siunitx}
\usepackage{amsmath}
\usepackage{enumitem}
\usepackage{caption}
\usepackage{subcaption}
\usepackage{float}
\usepackage{placeins}
\usepackage{xurl}
\usepackage[hidelinks]{hyperref}
\usepackage[nameinlink,noabbrev]{cleveref}
\usepackage{makecell}
\usepackage{longtable}
\usepackage{threeparttable}
\usepackage{pdflscape}

\hypersetup{
  pdftitle={Do Newer Lightweight CNNs Perform Better Under Resource Constraints?},
  pdfauthor={Tasnim Shahriar},
  pdfkeywords={lightweight convolutional neural networks, image classification, resource constraints, Pareto analysis, transfer learning, CPU latency, peak tensor memory, controlled benchmarking}
}

\graphicspath{{figures/}}
\setlist{nosep,leftmargin=*}
\newcolumntype{Y}{>{\centering\arraybackslash}X}
\title{Do Newer Lightweight CNNs Perform Better Under Resource Constraints?\\[4pt]
\large A Controlled Multigenerational Study of Architecture, Initialization, Training Budget, and Efficiency}
\author{Tasnim Shahriar\\{\itshape Independent Researcher}}
\date{}

\begin{document}
\maketitle

\begin{center}
\small\itshape This manuscript is currently under consideration at a peer reviewed venue. The preprint may be updated following feedback received during the review process.
\end{center}

\begin{abstract}
Newer lightweight convolutional neural network designs are often presented as offering improved predictive performance and deployment efficiency, but these advantages require evaluation under controlled downstream conditions. This study compares nine lightweight CNN model packages across CIFAR-10, CIFAR-100, and Tiny ImageNet using a shared training and evaluation protocol. Predictive performance is assessed using top-1 accuracy, macro F1, and top-5 accuracy, while resource demand is measured through parameter count, FP32 parameter storage, multiply-accumulate operations, standardized batch size 1 latency on an NVIDIA L4 and an AMD Ryzen 5 5500U CPU, and peak PyTorch CUDA allocated tensor memory. Accuracy and resource tradeoffs are further examined using independently calculated point-estimate Pareto frontiers. EfficientNetV2-S records the highest observed top-1 accuracy on CIFAR-10 and CIFAR-100 at 97.57\% and 86.98\%, whereas RepViT-M1.0 leads Tiny ImageNet at 79.87\%, exceeding EfficientNetV2-S by 1.14 percentage points. EfficientNet-B0 records 97.35\%, 86.13\%, and 78.08\% top-1 accuracy, remaining within 0.22, 0.85, and 1.79 percentage points of the best result on CIFAR-10, CIFAR-100, and Tiny ImageNet, respectively. Compared with EfficientNetV2-S, it uses approximately 79\% fewer parameters and 86\% fewer GMACs, while compared with RepViT-M1.0, it uses approximately 35\% fewer parameters and 64\% fewer GMACs. EfficientNet-B0 also belongs to each independently evaluated accuracy-resource Pareto frontier across all three datasets, supporting its role as the most consistently competitive intermediate-budget option. MobileNetV3-Small achieves 96.12\%, 82.43\%, and 70.95\% accuracy, records the lowest GMAC count, and is the fastest model under both evaluated CPU thread settings. MobileNetV4-Conv-S uses approximately 60\% more parameters and three times the GMACs of MobileNetV3-Small, yet records pretrained accuracy lower by 0.16, 0.53, and 1.04 percentage points across the three datasets. Under random initialization, MobileNetV3-Small achieves higher final test accuracy than MobileNetV4-Conv-S by 2.55, 1.76, and 0.99 percentage points, with paired test-set intervals excluding zero for the fixed trained models on all three datasets. The initialization study further shows that EfficientNet-B0 remains 3.29, 10.10, and 17.54 percentage points below its pretrained counterpart even after 100 epochs of scratch training, despite requiring approximately five times the recorded training time. At the smallest end of the model range, SqueezeNet1.1 has the lowest parameter count and peak CUDA allocation, but its substantially weaker predictive performance prevents it from matching the overall low-resource profile of MobileNetV3-Small. Latency rankings differ considerably between the evaluated L4 and CPU execution environments, showing that theoretical computation alone does not reliably predict measured inference performance. Overall, newer designs provide selective rather than universal gains. EfficientNetV2-S leads the CIFAR tasks, RepViT-M1.0 leads Tiny ImageNet, EfficientNet-B0 provides the most consistent general tradeoff, and MobileNetV3-Small emerges as the strongest evaluated option under severe resource constraints.
\end{abstract}


\section{Introduction}
Lightweight convolutional neural networks remain central to image classification on mobile, embedded, and resource-limited systems. Their design objective is rarely a single scalar quantity. A deployment may limit parameter storage, arithmetic cost, peak memory, latency, or training time, while still requiring an acceptable level of predictive quality. These quantities are related but not interchangeable. A model with few GMACs may execute slowly when its operators are fragmented or poorly supported, while a larger dense network may run efficiently on a highly parallel accelerator. Model selection under resource constraints is therefore a multiobjective problem rather than a simple accuracy leaderboard.

A second challenge is architectural turnover. New lightweight networks are typically introduced with architecture-specific pretraining, augmentation, distillation, input resolution, and hardware targets. Comparing their public checkpoints under one downstream recipe is practically useful, but it does not isolate architecture alone. This study therefore distinguishes two questions. The nine-model benchmark evaluates public pretrained architecture and checkpoint combinations under controlled downstream conditions. Separate scratch experiments provide narrower evidence about EfficientNet-B0 training budget and the MobileNetV3-Small versus MobileNetV4-Conv-S ordering.

The term \emph{controlled} refers to preprocessing, downstream optimization, input size, evaluation, and resource measurement. It does not imply that upstream pretraining is controlled. The term \emph{multigenerational} refers to a representative sample of design strategies from 2016 to 2024 rather than a complete chronology of every efficient CNN. The pool retains the seven established architectures from the earlier controlled benchmark and adds RepViT-M1.0 and MobileNetV4-Conv-S as two 2024 designs representing recent reparameterized and universal mobile-block directions. All nine backbones have public ImageNet-1K checkpoints, reproducible implementations, standard 224 pixel input support, fewer than approximately 25 million parameters, and distinct design mechanisms.

The study addresses five research questions:
\begin{itemize}
    \item \textbf{RQ1:} How do nine public pretrained model packages compare in top-1 accuracy, macro F1, and top-5 accuracy under one downstream protocol?
    \item \textbf{RQ2:} Which models remain point estimate Pareto efficient under parameters, GMACs, L4 latency, CPU latency, and peak CUDA tensor memory?
    \item \textbf{RQ3:} How strongly do theoretical resource indicators correspond to measured latency across the evaluated GPU and CPU execution environments?
    \item \textbf{RQ4:} How do separate 20 epoch and 100 epoch EfficientNet-B0 scratch schedules compare in accuracy and recorded wall clock time?
    \item \textbf{RQ5:} Does MobileNetV4-Conv-S improve upon MobileNetV3-Small under the evaluated pretrained and scratch settings?
\end{itemize}

The main contributions are:
\begin{itemize}
    \item A controlled downstream comparison of nine lightweight CNN model packages across three datasets, with explicit checkpoint provenance.
    \item A multidimensional resource evaluation using parameters, FP32 storage, GMACs, L4 latency distributions, consumer CPU latency, and peak CUDA allocated tensor memory.
    \item Point estimate Pareto frontiers, transparent example budgets, execution-environment rank analysis, and descriptive rank correlations that replace informal claims about a universal best balance.
    \item Focused training studies using cumulative wall clock time, extended EfficientNet-B0 scratch schedules, MobileNet cross-generation learning trajectories, and paired test-set uncertainty.
\end{itemize}

\section{Related Work}
\subsection{Compact CNN design and scaling}
Lightweight CNN research has progressed through several complementary strategies. SqueezeNet reduced parameter storage through Fire modules and aggressive use of $1\times1$ convolutions \cite{squeezenet}. ResNet established residual learning as a robust conventional baseline \cite{resnet}. MobileNetV2 introduced inverted residual blocks and linear bottlenecks \cite{mnv2}, while ShuffleNetV2 shifted attention from nominal FLOPs toward memory access cost, channel fragmentation, and practical operator design \cite{shuffle}. MobileNetV3 combined hardware-aware neural architecture search, squeeze-and-excitation, and specialized nonlinearities \cite{mnv3}. EfficientNet formalized compound scaling across depth, width, and resolution \cite{effnet}, and EfficientNetV2 introduced fused mobile blocks and training-aware scaling to improve both optimization and inference efficiency \cite{effnetv2}. Together, these works show that compactness can be pursued through parameter compression, block redesign, scaling policy, or hardware-aware search rather than one universal architectural recipe.

\subsection{Latency-aware and reparameterized design}
Recent efficient networks increasingly distinguish theoretical arithmetic from realized latency. MobileOne uses structural reparameterization so that a multi-branch training graph can be converted into a simpler inference graph \cite{mobileone}. FasterNet explicitly argues that low FLOPs do not guarantee fast execution and introduces partial convolution to reduce redundant memory access \cite{fasternet}. FastViT similarly combines structural reparameterization with hybrid convolutional blocks to improve latency and accuracy tradeoffs \cite{fastvit}. RepViT transfers efficient macro and micro design choices associated with mobile vision transformers back into a pure CNN \cite{repvit}. MobileNetV4 introduces the Universal Inverted Bottleneck, mobile multi-query attention, and a hardware-oriented search framework designed to remain competitive across CPUs, DSPs, GPUs, and accelerators \cite{mnv4}.

These studies motivate measuring parameters, arithmetic, memory, and latency separately. They also show why a model optimized for one device or runtime may not preserve the same ordering elsewhere. More recent compact CNNs continue to explore alternative mechanisms: StarNet studies multiplicative feature expansion \cite{starnet}, LSNet combines large-field perception with small-field aggregation \cite{lsnet}, and UniConvNet expands effective receptive fields while preserving the proposed spatial distribution of influence \cite{uniconvnet}. These models are relevant to the evolving design landscape, but they are not added to the experimental pool because the present study retains a fixed established benchmark base and adds only the two 2024 designs central to its cross-generation question.

\subsection{Backbone benchmarking, transfer, and constrained selection}
Large backbone studies demonstrate that rankings depend on the target task, upstream data, pretraining method, and adaptation procedure. Goldblum et al. compare a broad set of supervised, self-supervised, vision-language, and randomly initialized backbones across classification, detection, retrieval, and distribution-shift tasks \cite{goldblum}. Pegeot et al. examine compact-model transfer under upstream and downstream constraints, including linear probing and full fine tuning \cite{pegeot}. The RCV2023 challenges evaluate training and inference under explicit time, memory, and latency budgets \cite{tiwari}. Jeevan and Sethi compare pretrained backbones across multiple visual domains and reduced-data settings \cite{jeevan}. Guerin et al. formalize target-specific backbone selection in low-data regimes and show that no universal backbone dominates a large model pool \cite{guerin}.

This literature supports two distinctions used here. First, a shared downstream recipe provides protocol equality but does not isolate architecture from checkpoint provenance or guarantee architecture-specific optimality. Second, resource-aware selection requires explicit budgets or Pareto reasoning rather than an informal claim that one network has the best overall balance. The present analysis therefore treats each public pretrained architecture and checkpoint combination as a model package, and it calculates bivariate accuracy-resource frontiers independently for each resource dimension.

\subsection{Pretraining, optimization budget, and study positioning}
ImageNet pretraining often improves downstream accuracy and optimization efficiency, but the magnitude of the benefit depends on target data, task scale, and training schedule \cite{kornblith,rethinking}. Scratch and pretrained comparisons must therefore distinguish initialization from optimization exposure. Distillation further complicates public-checkpoint comparisons because it can improve a model package without changing the downstream architecture definition. These considerations motivate the separate EfficientNet-B0 schedule study and the matched MobileNet scratch comparison.

The pretrained results for the seven established architectures and the original fixed-budget EfficientNet-B0 initialization records were previously reported in \cite{shahriar2025}. RepViT-M1.0, MobileNetV4-Conv-S, the fresh 100 epoch schedules, unified latency and memory measurements, Pareto analyses, cumulative-time analyses, and paired prediction analyses are introduced in the present manuscript.

\begin{table}[H]
\centering
\caption{Scope of representative backbone and resource-aware studies relative to the present work.}
\label{tab:related}
\scriptsize
\begin{tabularx}{\textwidth}{p{2.5cm}p{3.7cm}p{3.7cm}X}
\toprule
\textbf{Study} & \textbf{Primary scope} & \textbf{Resource or adaptation focus} & \textbf{Scope relative to this study} \\
\midrule
Goldblum et al. \cite{goldblum} & Broad comparison of more than twenty backbones across multiple vision tasks & Pretraining paradigms, transfer, robustness, and random initialization & Much broader task and pretraining coverage; does not focus on a fixed multigenerational lightweight CNN pool with dual-environment latency and memory analysis \\
Pegeot et al. \cite{pegeot} & Transfer of compact architectures to six downstream tasks & Upstream data composition, linear probing, and full fine tuning under constraints & Emphasizes transfer mechanisms rather than cross-generation resource frontiers and extended scratch schedules \\
Tiwari et al. \cite{tiwari} & Resource-constrained training and inference challenges & Explicit time, memory, and latency budgets & Evaluates challenge systems rather than one shared-protocol architecture pool across three datasets \\
Jeevan and Sethi \cite{jeevan} & Domain-specific comparison of pretrained backbones & Reduced-data performance and resource-efficient fine tuning & Broader domain coverage; does not include the same CPU and GPU execution comparison or training-budget case studies \\
Guerin et al. \cite{guerin} & Target-specific backbone selection from a large model pool & Search cost and low-data model selection & Proposes a selection procedure rather than a fixed nine-model benchmark with detailed scratch analyses \\
Shahriar \cite{shahriar2025} & Established lightweight CNNs on the same three datasets & Predictive metrics, storage, GMACs, and initialization effects & Provides the established benchmark base; the current work adds recent models, new schedules, execution measurements, Pareto frontiers, and paired analysis \\
Present study & Nine CNN model packages spanning 2016 to 2024 & Accuracy, F1, top-5, static resources, GPU and CPU latency, peak CUDA memory, initialization, and recorded training time & Tests whether architectural recency yields consistent gains under one downstream protocol and several explicit resource definitions \\
\bottomrule
\end{tabularx}
\end{table}

\section{Experimental Methodology}
\subsection{Datasets and splits}
The benchmark uses CIFAR-10, CIFAR-100, and Tiny ImageNet. Ten percent of each official training set is reserved for validation through a fixed stratified split. The official CIFAR test sets are retained for final evaluation. Tiny ImageNet does not distribute labels for its official test set, so its labeled validation set is used as the benchmark test set, while the internal validation partition is drawn only from the original training images. Tiny ImageNet is derived from ImageNet and shares its class taxonomy, so it is not fully independent of ImageNet pretraining.

\begin{table}[H]
\centering
\caption{Dataset composition used in the experiments.}
\label{tab:datasets}
\begin{tabular}{lrrrr}
\toprule
\textbf{Dataset} & \textbf{Classes} & \textbf{Train} & \textbf{Validation} & \textbf{Test} \\
\midrule
CIFAR-10 & 10 & 45,000 & 5,000 & 10,000 \\
CIFAR-100 & 100 & 45,000 & 5,000 & 10,000 \\
Tiny ImageNet & 200 & 90,000 & 10,000 & 10,000 \\
\bottomrule
\end{tabular}
\end{table}

\subsection{Architectures and checkpoint provenance}
The seven established architectures are instantiated from torchvision using public ImageNet-1K weights distributed with the library. RepViT-M1.0 uses \texttt{repvit\_m1\_0.dist\_450e\_in1k} from timm. This checkpoint was trained with distillation by the RepViT authors. The timm implementation retains two classifier heads for distilled variants and averages their outputs during evaluation. The standardized 200-class RepViT therefore contains 6,584,396 parameters, and both heads are included consistently in parameter count, storage, GMACs, latency, and memory measurement. MobileNetV4-Conv-S uses \texttt{mobilenetv4\_conv\_small.e2400\_r224\_in1k} from timm, a public checkpoint trained with a MobileNetV4-inspired recipe rather than an official Google release. For every pretrained run, the source classifier is replaced for the target dataset and the full network is fine tuned.

\begin{table}[H]
\centering
\caption{Evaluated models and checkpoint provenance.}
\label{tab:models}
\scriptsize
\resizebox{\textwidth}{!}{%
\begin{tabular}{llllll}
\toprule
\textbf{Model} & \textbf{Year} & \textbf{Design family} & \textbf{Implementation} & \textbf{Public checkpoint} & \textbf{Provenance note} \\
\midrule
SqueezeNet1.1 & 2016 & Fire modules & torchvision & ImageNet-1K & torchvision-maintained ImageNet-1K weights \\
ResNet18 & 2016 & Residual blocks & torchvision & ImageNet-1K & torchvision-maintained ImageNet-1K weights \\
MobileNetV2 & 2018 & Inverted residuals & torchvision & ImageNet-1K & torchvision-maintained ImageNet-1K weights \\
ShuffleNetV2 x1.0 & 2018 & Channel split and shuffle & torchvision & ImageNet-1K & torchvision-maintained ImageNet-1K weights \\
EfficientNet-B0 & 2019 & Compound scaling & torchvision & ImageNet-1K & torchvision-maintained ImageNet-1K weights \\
MobileNetV3-Small & 2019 & Searched mobile blocks & torchvision & ImageNet-1K & torchvision-maintained ImageNet-1K weights \\
EfficientNetV2-S & 2021 & Fused mobile blocks & torchvision & ImageNet-1K & torchvision-maintained ImageNet-1K weights \\
RepViT-M1.0 & 2024 & ViT-inspired reparameterized CNN & timm 1.0.27 & \texttt{repvit\_m1\_0.dist\_450e\_in1k} & Distilled checkpoint, dual heads averaged at evaluation \\
MobileNetV4-Conv-S & 2024 & Universal Inverted Bottleneck & timm 1.0.27 & \texttt{mobilenetv4\_conv\_small.e2400\_r224\_in1k} & Public third-party checkpoint, not an official Google release \\
\bottomrule
\end{tabular}}
\end{table}

\subsection{Preprocessing and shared downstream protocol}
All images are processed at 224 by 224 pixels. Training uses random resized crop with scale 0.80 to 1.00 and aspect ratio 0.90 to 1.10, bicubic interpolation, horizontal flipping, color jitter, and ImageNet normalization. Validation and test images are resized deterministically and normalized with the same channel statistics. The 224 pixel setting preserves compatibility with source checkpoints, but it is a transfer adaptation choice rather than a native-resolution study. Upscaling does not create new image detail and increases computation relative to the original dataset resolution.

Models are optimized with AdamW using an initial learning rate of $3\times10^{-4}$, weight decay $10^{-4}$, PyTorch default coefficients $\beta=(0.9,0.999)$ and $\epsilon=10^{-8}$, label smoothing 0.1, mixed precision, and gradient clipping at 1.0. CosineAnnealingLR uses $T_{\max}$ equal to the configured schedule length and a minimum learning rate of $3\times10^{-6}$. Training batch size is 32 and evaluation batch size is 64. The common schedule has a maximum of 20 epochs with early stopping patience 5. Validation accuracy is the checkpoint criterion, with macro F1 used as a tie breaker. Training loaders shuffle with a seeded generator and drop the final incomplete batch; validation and test loaders use deterministic ordering. The training loader uses \texttt{pin\_memory=True}, seeded worker initialization, and persistent workers when worker processes are available. cuDNN deterministic mode is enabled and cuDNN benchmark mode is disabled. All reported training runs use seed 42 unless otherwise noted, and the primary benchmark reports one completed training run per setting.

\begin{table}[H]
\centering
\caption{Common downstream training configuration.}
\label{tab:training}
\begin{tabular}{llll}
\toprule
\textbf{Setting} & \textbf{Value} & \textbf{Setting} & \textbf{Value} \\
\midrule
Input resolution & 224 by 224 & Optimizer & AdamW \\
Maximum epochs & 20 & Initial learning rate & $3\times10^{-4}$ \\
Training batch size & 32 & Weight decay & $10^{-4}$ \\
Evaluation batch size & 64 & Label smoothing & 0.1 \\
Scheduler & Cosine annealing & Gradient clipping & 1.0 \\
Precision & Mixed precision & Checkpoint criterion & Validation accuracy \\
Early stopping & Patience 5 & Fine tuning & Full network \\
\bottomrule
\end{tabular}
\end{table}

\subsection{Scratch and training-budget experiments}
EfficientNet-B0 is evaluated under ImageNet pretrained initialization with the common schedule, random initialization with the same maximum 20 epoch schedule, and a fresh random initialization run using a cosine schedule configured for 100 epochs. The 100 epoch runs start from a new random initialization rather than continuing the completed 20 epoch schedule. They use early stopping patience 15. Comparisons therefore describe two distinct schedules, not the effect of simply appending 80 epochs.

MobileNetV3-Small and MobileNetV4-Conv-S are evaluated under the public pretrained protocol and under matched 20 epoch scratch training. These experiments measure learning behavior under one fixed downstream budget. They do not estimate architecture-specific maximum scratch performance.

All recorded training-time analyses use the NVIDIA L4 environment reported in the saved experiment metadata. The main software environment used Python 3.12.13, PyTorch 2.11.0 with CUDA 12.8, torchvision 0.26.0, and timm 1.0.27 for the recent models.

\subsection{Resource, latency, memory, and Pareto measurement}
Static efficiency is described by total parameters, FP32 parameter storage, and GMACs for input shape $1\times3\times224\times224$. Cross-model resource measurements use standardized 200-class variants so that one classifier size is applied consistently. These figures are standardized proxies for comparing architecture instances and are not the exact CIFAR-10 classifier footprints.

NVIDIA L4 latency uses PyTorch eager execution, FP32 model parameters, FP16 CUDA autocast, batch size 1, 50 warmups, and 500 timed forward passes measured with CUDA Events. Host transfer and preprocessing are excluded. The same run records baseline allocated tensor memory, peak allocated tensor memory, incremental inference peak above the loaded-model baseline, and allocator-reserved memory. The main memory metric is peak PyTorch CUDA allocated tensor memory, not total device memory.

Consumer CPU latency is measured on an AMD Ryzen 5 5500U using Windows 11, PyTorch 2.12.1 CPU, FP32, MKLDNN, batch size 1, and contiguous NCHW input. One-thread and four-thread modes are evaluated in fresh subprocesses. Each configuration uses 30 warmups followed by three rounds of 100 measured passes. Median and p95 latency are emphasized because occasional operating-system activity can affect tail measurements.

For each dataset and resource metric, a model is point estimate Pareto optimal when no other evaluated model is both at least as accurate and no more resource demanding, with one strict improvement, following the standard bivariate Pareto concept in multiobjective optimization \cite{miettinen}. Each frontier is calculated independently using top-1 accuracy and one resource dimension at a time; the study does not construct one simultaneous multidimensional frontier. These frontiers do not include training-seed uncertainty, so membership near small accuracy gaps may change after retraining.

\subsection{Paired test-set analysis}
For selected model pairs, the stored predictions are compared over the same ordered test examples. Accuracy differences are summarized with 50,000 paired bootstrap replicates \cite{efron} and exact two-sided McNemar tests \cite{mcnemar}. The prediction files do not contain explicit sample identifiers, but true-label sequences match row by row and the evaluation loaders use deterministic ordering. The comparisons are exploratory and are not adjusted for multiplicity; they are used to qualify selected observed differences rather than support confirmatory architecture-level hypothesis testing. This analysis quantifies uncertainty over fixed test examples for the trained models. It does not capture variability across independent training runs.

\section{Results}
\subsection{Predictive performance of pretrained model packages}
Table~\ref{tab:predictive} reports top-1 accuracy and macro F1. EfficientNetV2-S ranks first on CIFAR-10 and CIFAR-100. RepViT-M1.0 ranks third on both CIFAR datasets but leads Tiny ImageNet at 79.87\%, exceeding EfficientNetV2-S by 1.14 points and EfficientNet-B0 by 1.79 points. MobileNetV3-Small remains competitive despite its compact resource profile. MobileNetV4-Conv-S records lower observed top-1 accuracy than MobileNetV3-Small by 0.16, 0.53, and 1.04 points across the three datasets.

Macro F1 closely follows top-1 accuracy because the evaluation datasets are class balanced. Top-5 accuracy provides additional information on the higher-class-count tasks. EfficientNet-B0 leads CIFAR-100 top-5 accuracy at 97.29\%, while RepViT-M1.0 leads Tiny ImageNet at 93.09\%.

\begin{table}[H]
\centering
\caption{Pretrained top-1 accuracy and macro F1. Best values are bold and second-best values are underlined.}
\label{tab:predictive}
\scriptsize
\begin{threeparttable}
\resizebox{\textwidth}{!}{%
\begin{tabular}{lrrrrrr}
\toprule
& \multicolumn{2}{c}{\textbf{CIFAR-10}} & \multicolumn{2}{c}{\textbf{CIFAR-100}} & \multicolumn{2}{c}{\textbf{Tiny ImageNet}} \\
\cmidrule(lr){2-3}\cmidrule(lr){4-5}\cmidrule(lr){6-7}
\textbf{Model} & \textbf{Acc. (\%)} & \textbf{Macro F1} & \textbf{Acc. (\%)} & \textbf{Macro F1} & \textbf{Acc. (\%)} & \textbf{Macro F1} \\
\midrule
EfficientNetV2-S & \textbf{97.57} & \textbf{0.9757} & \textbf{86.98} & \textbf{0.8697} & \underline{78.73} & \underline{0.7866} \\
EfficientNet-B0 & \underline{97.35} & \underline{0.9735} & \underline{86.13} & \underline{0.8615} & 78.08 & 0.7801 \\
RepViT-M1.0 & 97.25 & 0.9725 & 85.40 & 0.8534 & \textbf{79.87} & \textbf{0.7981} \\
ResNet18 & 96.52 & 0.9652 & 82.53 & 0.8250 & 68.22 & 0.6815 \\
MobileNetV3-Small & 96.12 & 0.9611 & 82.43 & 0.8239 & 70.95 & 0.7093 \\
MobileNetV2 & 95.99 & 0.9599 & 81.45 & 0.8144 & 71.41 & 0.7126 \\
MobileNetV4-Conv-S & 95.96 & 0.9595 & 81.90 & 0.8183 & 69.91 & 0.6979 \\
ShuffleNetV2 x1.0 & 94.78 & 0.9477 & 80.38 & 0.8036 & 69.35 & 0.6925 \\
SqueezeNet1.1 & 93.73 & 0.9371 & 73.34 & 0.7330 & 60.41 & 0.6030 \\
\bottomrule
\end{tabular}}
\begin{tablenotes}\footnotesize
\item The pretrained results for the seven established architectures and the original fixed-budget EfficientNet-B0 initialization records were previously reported in \cite{shahriar2025}. RepViT-M1.0, MobileNetV4-Conv-S, the fresh 100 epoch schedules, unified latency and memory measurements, Pareto analyses, cumulative-time analyses, and paired prediction analyses are new to this manuscript.
\end{tablenotes}
\end{threeparttable}
\end{table}

\begin{figure}[H]
\centering
\includegraphics[width=\textwidth]{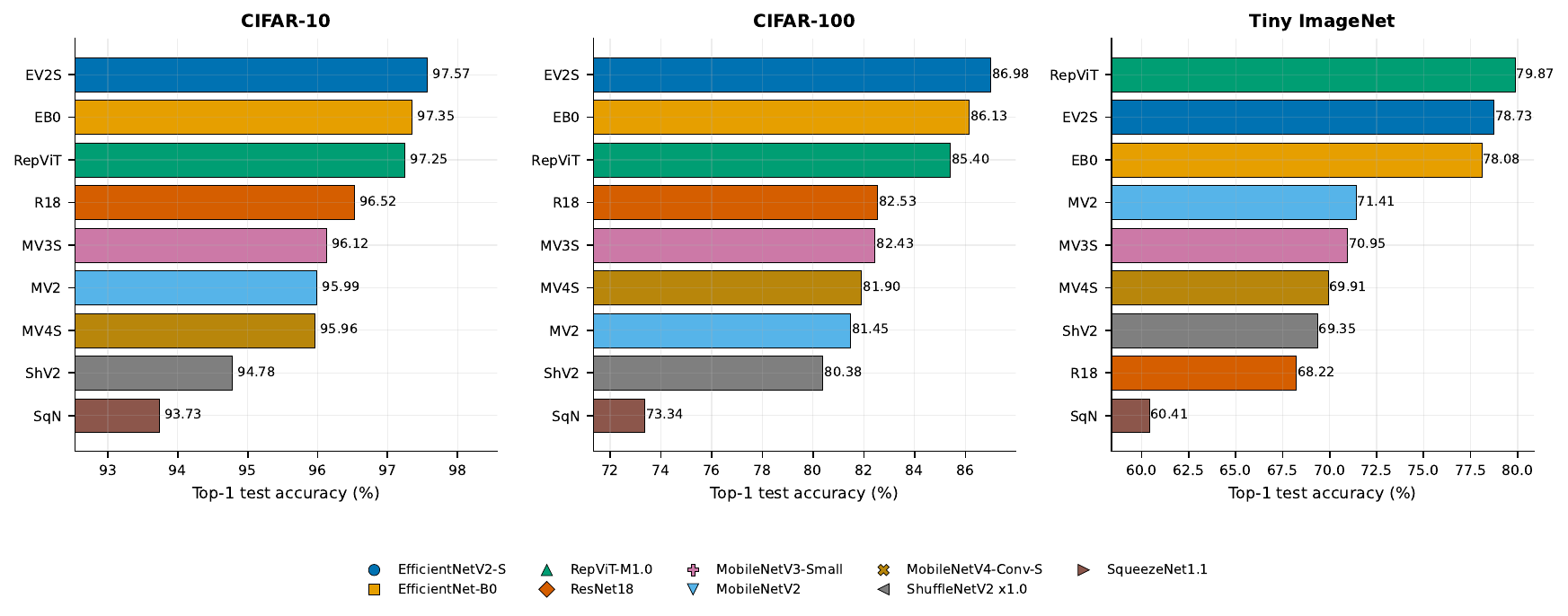}
\caption{Pretrained top-1 test accuracy. Each model uses the same color and marker identity throughout the paper.}
\label{fig:accuracy}
\end{figure}

Full top-5 results are provided in Appendix~\ref{app:top5}.

\subsection{Static resources, peak memory, and point estimate Pareto frontiers}
Table~\ref{tab:resources} reports standardized 200-class resource measurements. EfficientNetV2-S is the largest and most computationally demanding model. EfficientNet-B0 uses approximately 79\% fewer parameters and 86\% fewer GMACs than EfficientNetV2-S while trailing it by only 0.22 and 0.85 points on the CIFAR tasks. Relative to the Tiny ImageNet leader, RepViT-M1.0, EfficientNet-B0 uses approximately 35\% fewer parameters and 64\% fewer GMACs while trailing by 1.79 points.

MobileNetV3-Small defines the lowest arithmetic budget at 0.0617 GMACs. It uses 37\% fewer parameters and 67\% fewer GMACs than MobileNetV4-Conv-S, while recording higher observed accuracy on all three datasets. SqueezeNet1.1 has the lowest parameter count and the smallest peak CUDA allocation, but its substantially lower CIFAR-100 and Tiny ImageNet accuracy limits its practical profile.

\begin{table}[H]
\centering
\caption{Static resource measures and peak CUDA tensor memory for standardized 200-class variants.}
\label{tab:resources}
\scriptsize
\resizebox{\textwidth}{!}{%
\begin{tabular}{lrrrrr}
\toprule
\textbf{Model} & \textbf{Params (M)} & \textbf{FP32 (MiB)} & \textbf{GMACs} & \textbf{Peak CUDA allocated (MiB)} & \textbf{Incremental inference peak (MiB)} \\
\midrule
EfficientNetV2-S & 20.434 & 77.948 & 2.9009 & 93.11 & 3.76 \\
EfficientNet-B0 & 4.264 & 16.265 & 0.4141 & 31.18 & 4.98 \\
RepViT-M1.0 & 6.584 & 25.117 & 1.1420 & 37.74 & 2.57 \\
ResNet18 & 11.279 & 43.026 & 1.8236 & 62.24 & 9.46 \\
MobileNetV3-Small & 1.723 & 6.572 & \textbf{0.0617} & 17.51 & \textbf{1.13} \\
MobileNetV2 & 2.480 & 9.461 & 0.3265 & 24.34 & 4.98 \\
MobileNetV4-Conv-S & 2.749 & 10.487 & 0.1848 & 22.84 & 2.51 \\
ShuffleNetV2 x1.0 & 1.459 & 5.564 & 0.1519 & 17.79 & 2.39 \\
SqueezeNet1.1 & \textbf{0.825} & \textbf{3.147} & 0.2800 & \textbf{7.08} & 3.35 \\
\bottomrule
\end{tabular}}
\end{table}

\begin{figure}[H]
\centering
\includegraphics[width=\textwidth]{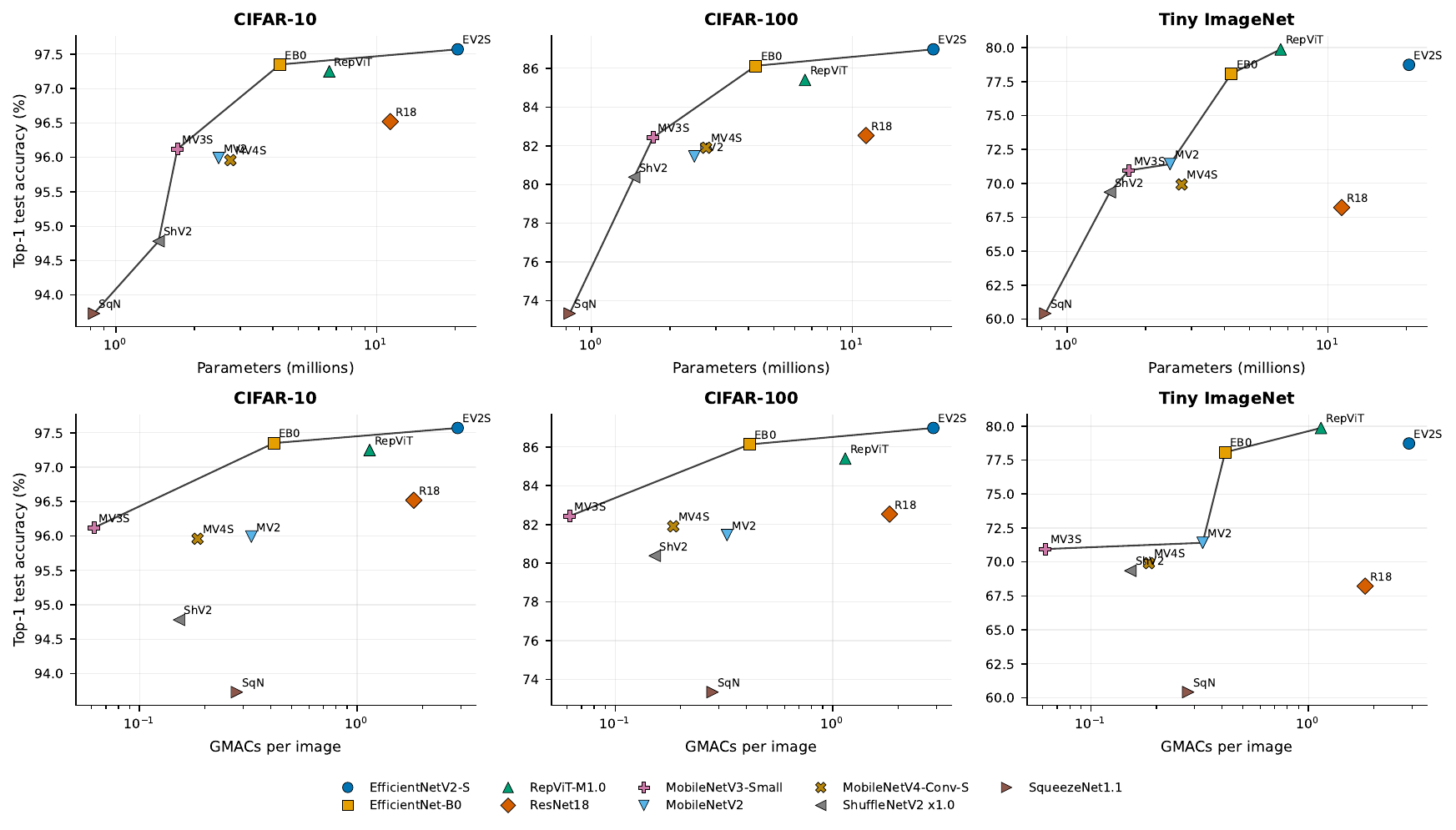}
\caption{Top-1 accuracy against parameters and GMACs. Black lines connect point estimate Pareto-efficient models in order of increasing resource demand.}
\label{fig:staticpareto}
\end{figure}

Across CIFAR-10 and CIFAR-100, the parameter frontier contains SqueezeNet1.1, ShuffleNetV2 x1.0, MobileNetV3-Small, EfficientNet-B0, and EfficientNetV2-S. The GMAC frontier contains MobileNetV3-Small, EfficientNet-B0, and EfficientNetV2-S. Tiny ImageNet changes the ordering: MobileNetV2 and RepViT-M1.0 become important frontier members, reflecting the stronger performance of these models on the more difficult 200-class task.

Peak allocated memory is strongly associated with parameter count in this model pool, with Spearman $\rho=0.97$. It nevertheless captures activation, buffer, output, and temporary tensor demand that FP32 parameter storage alone does not. Appendix Figure~\ref{fig:peakmemory} shows that the peak-memory frontiers broadly mirror the parameter frontiers, while ResNet18 has the largest incremental inference peak at 9.46 MiB.

The corresponding accuracy versus peak-memory plots are provided in Appendix~\ref{app:memoryplot}.

\subsection{Measured latency and execution-environment-specific ranking}
Table~\ref{tab:latency} reports median and p95 latency. ResNet18 is the fastest L4 model at 3.038 ms median latency, narrowly ahead of SqueezeNet1.1. MobileNetV3-Small is the fastest CPU model in both thread modes at 14.20 and 14.25 ms. MobileNetV4-Conv-S is approximately 59\% slower than MobileNetV3-Small in one-thread median latency and approximately 32\% slower in four-thread median latency.

RepViT-M1.0 scales strongly from 199.80 ms with one thread to 61.78 ms with four threads, but remains substantially slower than MobileNetV3-Small and EfficientNet-B0 on the evaluated CPU. SqueezeNet1.1 and ResNet18 also benefit substantially from four-thread execution. These results show that parallel scaling and operator support differ materially across architectures.

\begin{table}[H]
\centering
\caption{Batch size 1 measured latency. L4 uses FP16 autocast. CPU uses FP32. Preprocessing and host transfer are excluded.}
\label{tab:latency}
\scriptsize
\resizebox{\textwidth}{!}{%
\begin{tabular}{lrrrrrr}
\toprule
\textbf{Model} & \multicolumn{2}{c}{\textbf{NVIDIA L4 (ms)}} & \multicolumn{2}{c}{\textbf{CPU, 1 thread (ms)}} & \multicolumn{2}{c}{\textbf{CPU, 4 threads (ms)}} \\
\cmidrule(lr){2-3}\cmidrule(lr){4-5}\cmidrule(lr){6-7}
& \textbf{Median} & \textbf{p95} & \textbf{Median} & \textbf{p95} & \textbf{Median} & \textbf{p95} \\
\midrule
EfficientNetV2-S & 22.086 & 22.860 & 213.12 & 215.95 & 112.94 & 131.57 \\
EfficientNet-B0 & 9.832 & 10.000 & 69.50 & 77.05 & 43.24 & 49.65 \\
RepViT-M1.0 & 16.113 & 16.766 & 199.80 & 202.26 & 61.78 & 68.54 \\
ResNet18 & \textbf{3.038} & \textbf{3.154} & 127.33 & 129.05 & 46.95 & 58.63 \\
MobileNetV3-Small & 6.444 & 6.594 & \textbf{14.20} & \textbf{14.64} & \textbf{14.25} & \textbf{14.90} \\
MobileNetV2 & 6.437 & 6.564 & 41.83 & 42.87 & 31.13 & 45.32 \\
MobileNetV4-Conv-S & 6.389 & 6.707 & 22.58 & 23.12 & 18.76 & 20.06 \\
ShuffleNetV2 x1.0 & 7.554 & 7.694 & 34.21 & 35.15 & 25.43 & 29.15 \\
SqueezeNet1.1 & 3.054 & 3.196 & 67.19 & 81.75 & 23.97 & 27.08 \\
\bottomrule
\end{tabular}}
\end{table}

\begin{figure}[H]
\centering
\includegraphics[width=\textwidth]{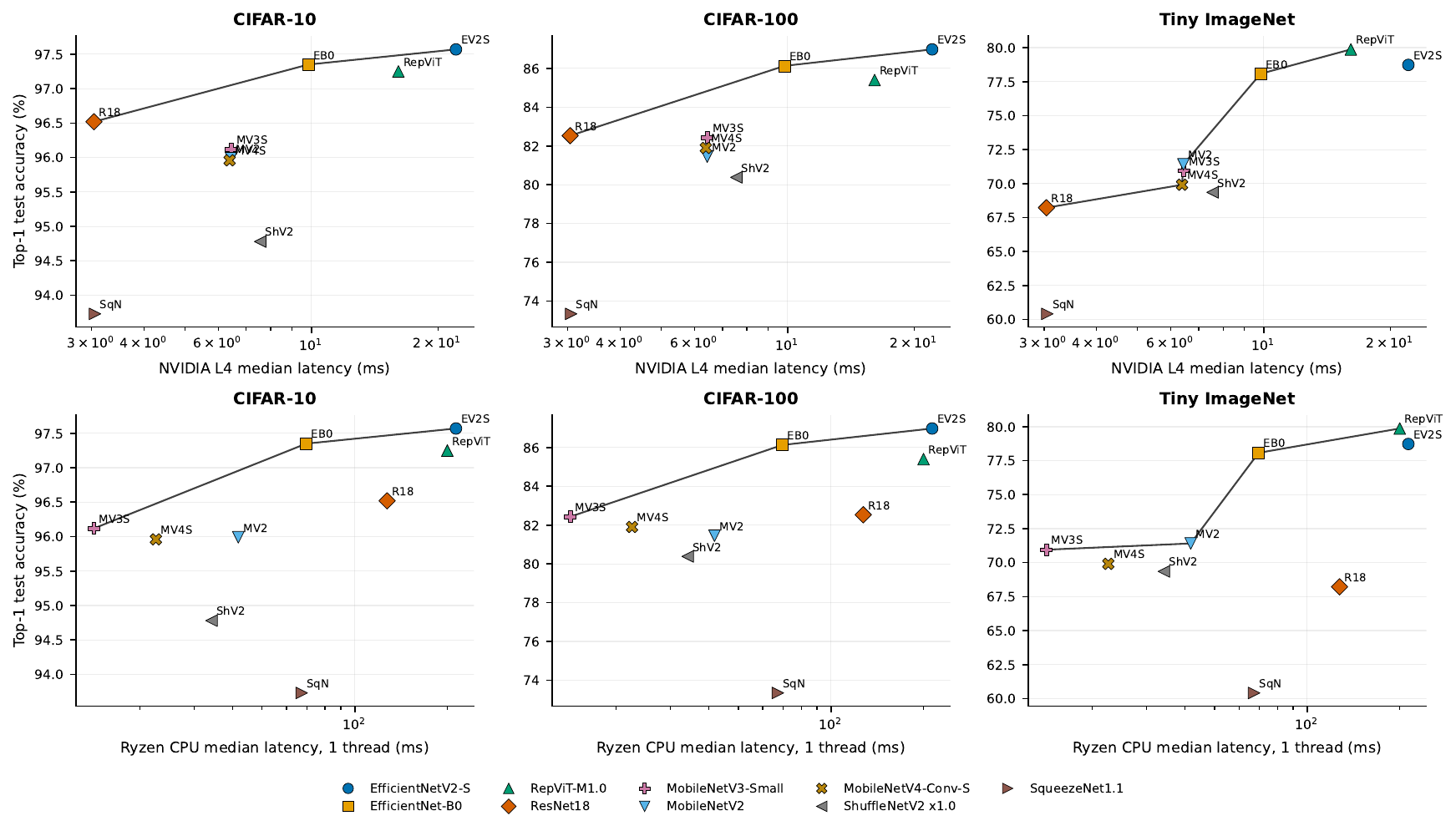}
\caption{Top-1 accuracy against NVIDIA L4 median latency and Ryzen CPU one-thread median latency. Point estimate Pareto frontiers differ between the evaluated execution environments.}
\label{fig:latencypareto}
\end{figure}

\begin{figure}[H]
\centering
\includegraphics[width=0.82\textwidth]{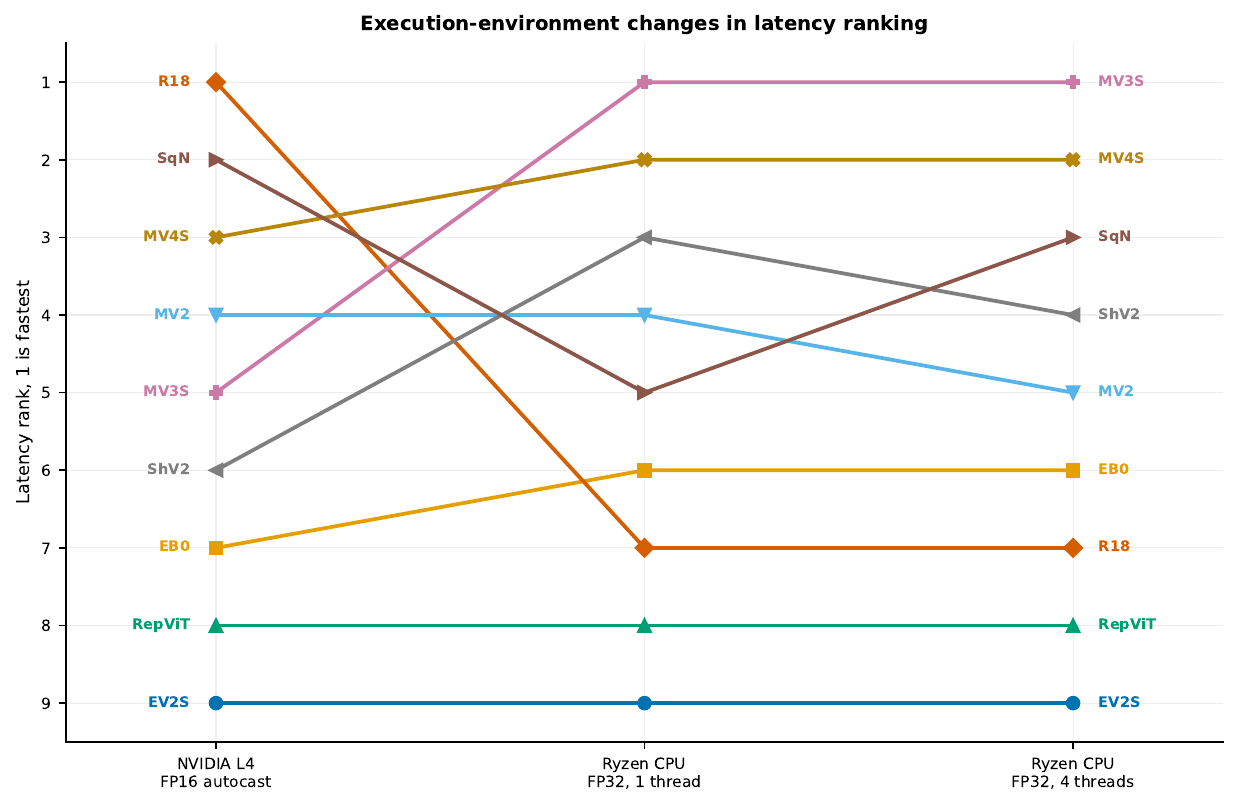}
\caption{Latency rank changes across the evaluated NVIDIA L4 FP16 autocast and AMD Ryzen 5 5500U FP32 environments.}
\label{fig:rankshift}
\end{figure}

The execution-environment-specific ranking shift is substantial. ResNet18 ranks first on the L4 but seventh in one-thread CPU latency. MobileNetV3-Small ranks fifth on the L4 but first under both CPU thread modes. SqueezeNet1.1 ranks second on the L4, fifth with one CPU thread, and third with four CPU threads. Figure~\ref{fig:rankshift} visualizes these changes directly.

Descriptive Spearman rank correlations \cite{spearman} reinforce the same conclusion. GMACs correlate strongly with CPU median latency, $\rho=0.95$ for one thread and $\rho=0.93$ for four threads, but weakly with L4 median latency, $\rho=0.27$. L4 and one-thread CPU latency rankings correlate only moderately, $\rho=0.40$. These values describe the nine-model sample and should not be generalized as population estimates.

The complete descriptive correlation table is provided in Appendix~\ref{app:correlations}.

\subsection{Illustrative resource budgets}
Point estimate Pareto frontiers describe all nondominated options but do not select one model without a deployment budget. Table~\ref{tab:profiles} gives transparent illustrative profiles. The thresholds are examples derived from natural breakpoints in the evaluated model pool, not universal deployment standards.

\begin{table}[H]
\centering
\caption{Illustrative resource profiles and highest observed feasible model.}
\label{tab:profiles}
\scriptsize
\begin{tabularx}{\textwidth}{p{3.0cm}p{4.2cm}p{4.0cm}X}
\toprule
\textbf{Profile} & \textbf{Budget} & \textbf{CIFAR-10 and CIFAR-100} & \textbf{Tiny ImageNet} \\
\midrule
Very compact storage & At most 2M parameters & MobileNetV3-Small & MobileNetV3-Small \\
Severe arithmetic limit & At most 0.10 GMAC & MobileNetV3-Small & MobileNetV3-Small \\
Compact general deployment & At most 5M parameters and 0.50 GMAC & EfficientNet-B0 & EfficientNet-B0 \\
Low CPU latency & At most 25 ms, one thread & MobileNetV3-Small & MobileNetV3-Small \\
Accuracy-oriented deployment & No strict compact budget & EfficientNetV2-S & RepViT-M1.0 \\
\bottomrule
\end{tabularx}
\end{table}

EfficientNet-B0 is notable because it belongs to each independently calculated bivariate accuracy-resource frontier for all six resource dimensions: parameters, GMACs, L4 latency, one-thread CPU latency, four-thread CPU latency, and peak CUDA allocated memory. This does not make it a universal winner, but it supports its role as the most consistently competitive intermediate-budget option in the evaluated pool.

\subsection{EfficientNet-B0 initialization and training budget}
Under the common schedule, ImageNet pretraining improves EfficientNet-B0 test accuracy by 5.38 points on CIFAR-10, 14.90 points on CIFAR-100, and 20.65 points on Tiny ImageNet. A fresh 100 epoch scratch schedule increases scratch accuracy by 2.09, 4.80, and 3.11 points relative to the separate 20 epoch schedule, but the pretrained model still leads by 3.29, 10.10, and 17.54 points.

The longer schedules require approximately five times the recorded training time: 159.65 versus 31.59 minutes on CIFAR-10, 154.31 versus 31.55 minutes on CIFAR-100, and 314.52 versus 63.55 minutes on Tiny ImageNet. The gains are therefore meaningful but expensive, especially on the higher-class-count datasets.

\begin{table}[H]
\centering
\caption{EfficientNet-B0 initialization and training-budget results. Training time is recorded on the NVIDIA L4.}
\label{tab:effbudget}
\scriptsize
\resizebox{\textwidth}{!}{%
\begin{tabular}{lrrrrrrr}
\toprule
\textbf{Dataset} & \textbf{Pretrained test} & \textbf{Scratch 20 test} & \textbf{Scratch 100 test} & \textbf{Remaining gap} & \textbf{20 epoch time (min)} & \textbf{100 epoch time (min)} & \textbf{Time multiplier} \\
\midrule
CIFAR-10 & 97.35 & 91.97 & 94.06 & 3.29 & 31.59 & 159.65 & 5.06 \\
CIFAR-100 & 86.13 & 71.23 & 76.03 & 10.10 & 31.55 & 154.31 & 4.89 \\
Tiny ImageNet & 78.08 & 57.43 & 60.54 & 17.54 & 63.55 & 314.52 & 4.95 \\
\bottomrule
\end{tabular}}
\end{table}

\begin{figure}[H]
\centering
\includegraphics[width=\textwidth]{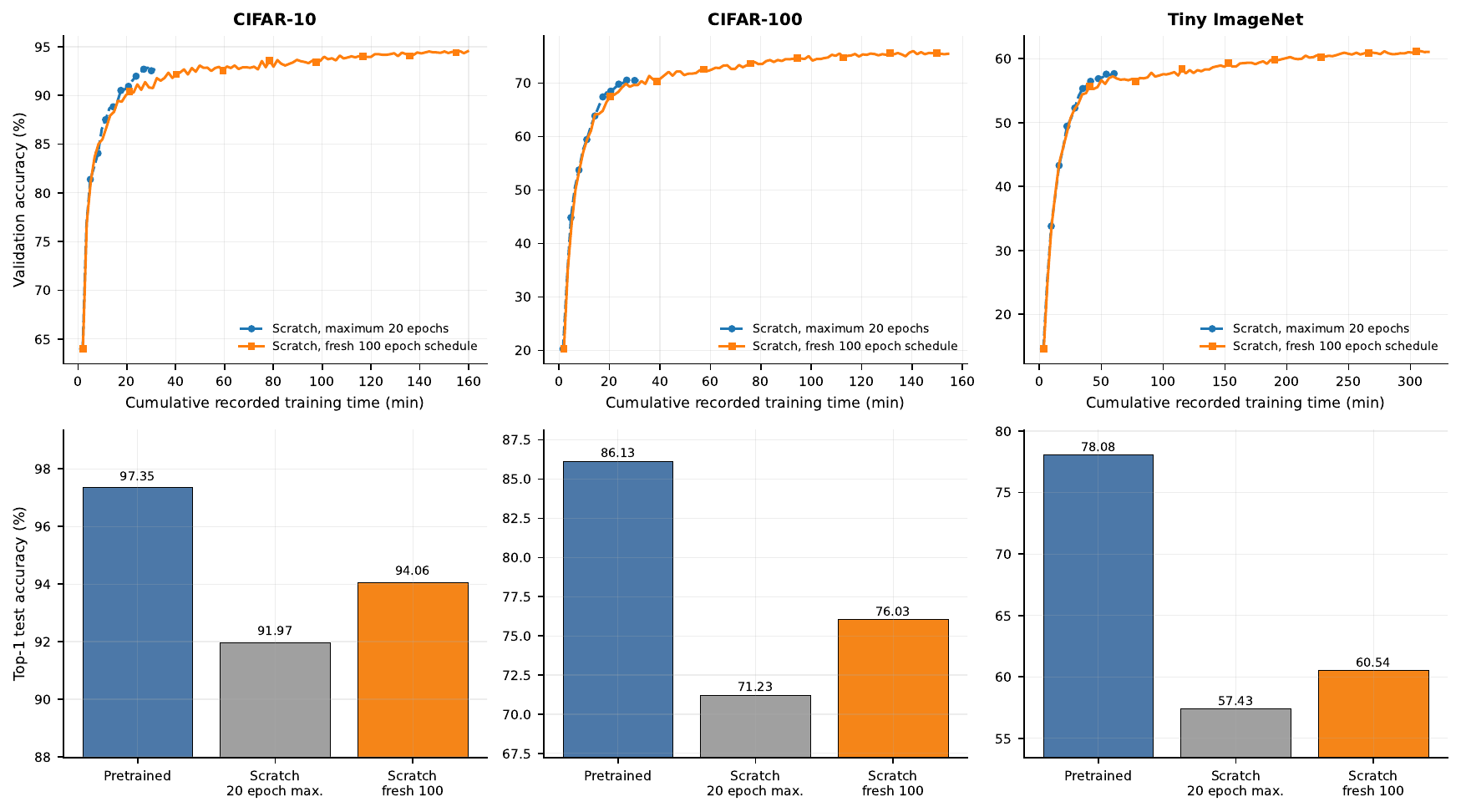}
\caption{EfficientNet-B0 scratch validation accuracy against cumulative recorded training time, with separate endpoint bars for pretrained, maximum 20 epoch scratch, and fresh 100 epoch scratch schedules. The scratch curves represent distinct runs rather than one continued schedule.}
\label{fig:effbudget}
\end{figure}

\subsection{MobileNetV3-Small and MobileNetV4-Conv-S}
Under pretrained initialization, MobileNetV3-Small records higher observed test accuracy than MobileNetV4-Conv-S by 0.16 points on CIFAR-10, 0.53 points on CIFAR-100, and 1.04 points on Tiny ImageNet. The paired intervals cross zero for the two CIFAR differences, while the Tiny ImageNet interval is 0.23 to 1.84 points.

Under scratch training, MobileNetV3-Small leads by 2.55, 1.76, and 0.99 points. The paired intervals are 1.90 to 3.21, 0.86 to 2.67, and 0.14 to 1.84 points, with exact McNemar $p<0.05$ for all three datasets. The learning curves show different behavior across tasks. MobileNetV3-Small leads throughout CIFAR-10, overtakes MobileNetV4-Conv-S on CIFAR-100, and remains close but ends higher on Tiny ImageNet. The cumulative-time curves do not reverse the final ordering.

\begin{figure}[H]
\centering
\includegraphics[width=\textwidth]{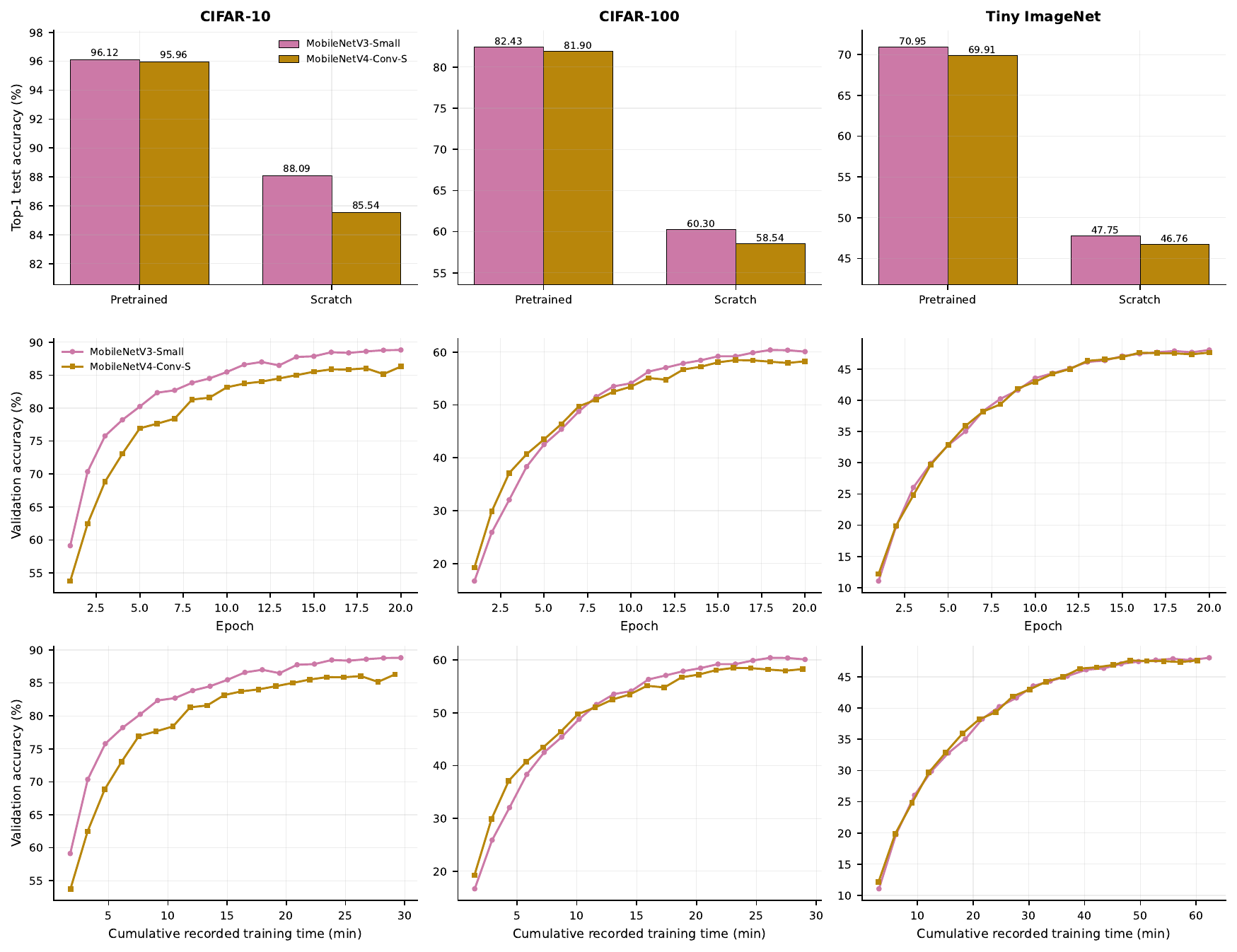}
\caption{MobileNetV3-Small and MobileNetV4-Conv-S final accuracy, scratch validation accuracy by epoch, and scratch validation accuracy by cumulative recorded training time.}
\label{fig:mobile}
\end{figure}

The result supports a conditional conclusion. The evaluated later-generation MobileNetV4-Conv-S model package does not improve upon MobileNetV3-Small under the tested public checkpoints, downstream recipe, scratch budget, CPU environment, parameter budget, or GMAC budget. It does not establish that MobileNetV4 is universally inferior under architecture-specific training or on its intended deployment targets.

\subsection{Paired test-set uncertainty}
\begin{figure}[H]
\centering
\includegraphics[width=0.92\textwidth]{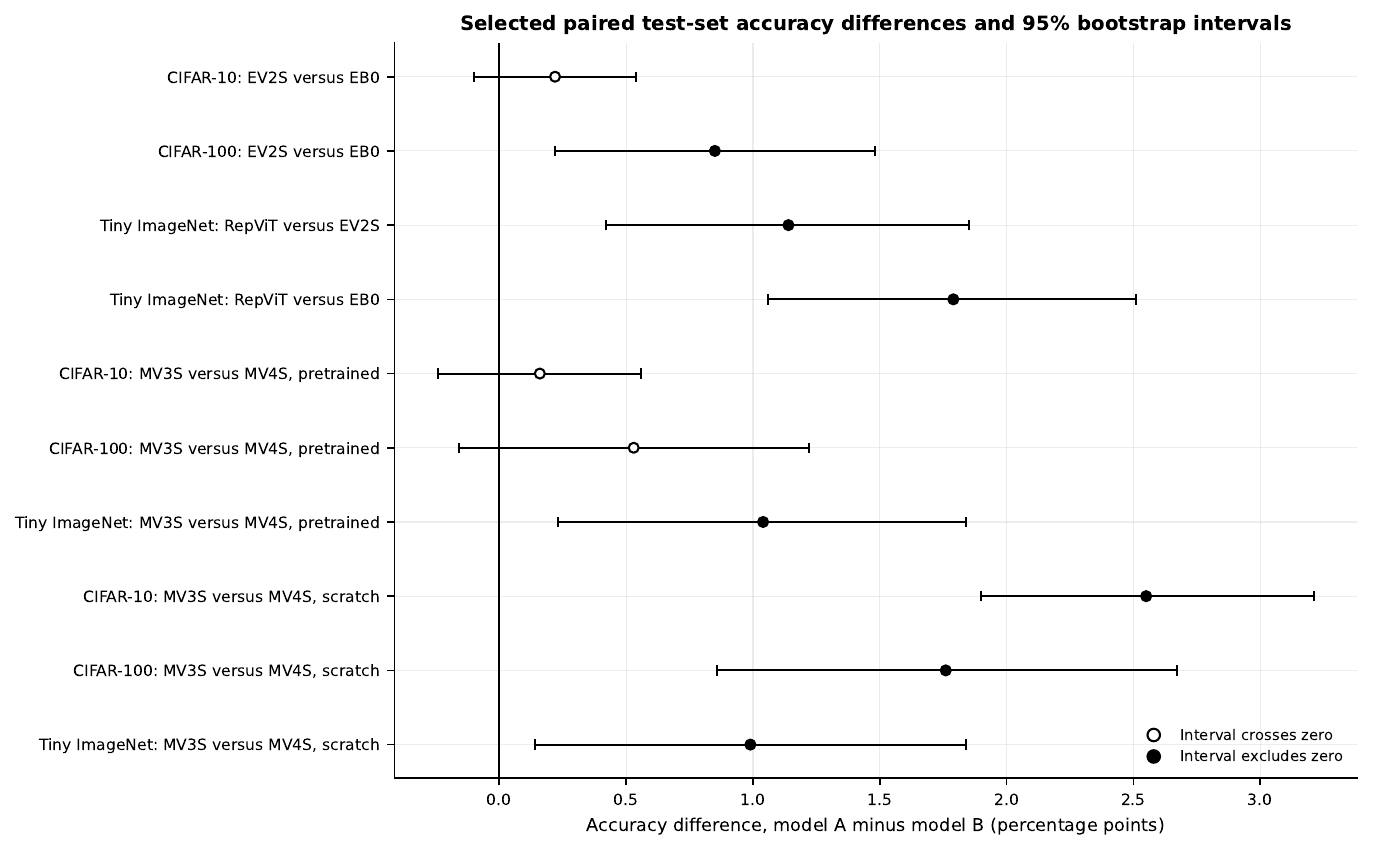}
\caption{Selected paired test-set accuracy differences and 95\% bootstrap intervals. Open markers indicate intervals that cross zero, while filled markers indicate intervals that exclude zero. These intervals condition on one trained model per setting and do not capture training-seed variability.}
\label{fig:paired}
\end{figure}

Figure~\ref{fig:paired} summarizes selected paired accuracy differences. Several small pretrained differences are not clearly separated from zero, including EfficientNetV2-S versus EfficientNet-B0 on CIFAR-10, EfficientNet-B0 versus RepViT-M1.0 on CIFAR-10, and MobileNetV3-Small versus MobileNetV4-Conv-S on both CIFAR datasets. In contrast, RepViT's Tiny ImageNet advantages and the three MobileNet scratch differences are supported by intervals that exclude zero.

The numerical paired-comparison table is provided in Appendix~\ref{app:pairedtable}.

\section{Discussion}
\subsection{Do newer lightweight CNNs perform better?}
The results do not support a monotonic relationship between release year and downstream performance. RepViT-M1.0 provides a clear selective gain on Tiny ImageNet, where it leads top-1 accuracy, macro F1, and top-5 accuracy. EfficientNetV2-S remains the highest observed CIFAR model. MobileNetV4-Conv-S provides the opposite result: it does not exceed MobileNetV3-Small under the evaluated pretrained or scratch conditions and is less favorable in parameters, GMACs, peak CUDA memory, and CPU latency.

Accuracy rankings are highly consistent between CIFAR-10 and CIFAR-100, with Spearman $\rho=0.98$, but less consistent between either CIFAR dataset and Tiny ImageNet, with $\rho=0.77$ and $\rho=0.73$. This supports the view that later designs may improve a particular target task without becoming universal winners.

\subsection{Model profiles under different constraints}
The benchmark supports several model profiles rather than one universal ranking. EfficientNetV2-S is the accuracy-oriented choice for the CIFAR tasks when its higher resource demand is acceptable. RepViT-M1.0 is the accuracy-oriented Tiny ImageNet choice. EfficientNet-B0 is the most consistently Pareto-efficient intermediate option. It remains close to the dataset leaders while belonging to each independently calculated bivariate accuracy-resource frontier.

MobileNetV3-Small provides the highest observed accuracy under the evaluated severe arithmetic and one-thread CPU budgets. It has the lowest GMAC count, leads both CPU latency modes, and records substantially stronger accuracy than the smallest-storage SqueezeNet1.1. SqueezeNet remains useful when parameter storage or peak CUDA allocation is the dominant constraint, but its predictive losses become large on CIFAR-100 and Tiny ImageNet.

\subsection{Why theoretical and measured efficiency disagree}
Parameters, GMACs, latency, and peak tensor memory answer different deployment questions. GMACs correspond strongly to latency on the tested CPU, but poorly to L4 latency. Dense conventional operations in ResNet18 execute efficiently on the L4 despite high arithmetic demand. MobileNetV3-Small shows the reverse execution-environment shift: it is not the fastest L4 model, but it decisively leads the CPU benchmark. Operator fusion, memory access, kernel selection, parallelism, and thread overhead therefore influence realized latency in addition to nominal computation.

The execution-environment comparison should remain narrow. The findings describe one NVIDIA L4 FP16 autocast environment and one AMD Ryzen 5 5500U FP32 environment. They do not establish rankings on mobile phones, ARM boards, neural processing units, DSPs, or production deployment runtimes.

\subsection{Pretraining and optimization budget}
The EfficientNet-B0 results separate initialization from training duration. Pretraining yields large advantages under the common schedule. The fresh 100 epoch scratch schedules produce higher observed accuracy than the separate shorter schedules, but the remaining gaps are larger on CIFAR-100 and Tiny ImageNet. The approximately fivefold increase in recorded training time also shows why equal epoch counts and equal computational exposure are not equivalent concepts.

The MobileNet experiment addresses a different question: comparative learning under one constrained schedule. MobileNetV3-Small reaches stronger final scratch performance and generally stronger learning trajectories. This reduces, but does not eliminate, checkpoint provenance as an explanation for the pretrained ordering.

\subsection{Interpreting the paired analysis}
Paired test-set analysis prevents small point differences from being overstated. The CIFAR-10 EfficientNetV2-S versus EfficientNet-B0 difference and the CIFAR pretrained MobileNet differences are not clearly separated by finite test-set uncertainty. The Tiny ImageNet RepViT advantages and all three MobileNet scratch advantages are better supported. However, these results condition on one completed training run and cannot estimate the variance produced by independent initialization, data order, or nondeterministic operations.

\section{Limitations}
The principal limitation is the absence of repeated training seeds. All predictive results are point estimates from one completed training seed, and paired test-set intervals do not substitute for training-run uncertainty. Close differences should therefore be interpreted cautiously.

The shared recipe provides protocol equality rather than architecture-specific opportunity equality. The separate EfficientNet-B0 scratch schedules use the same nominal seed but are not a matched-initial-state intervention; they differ in schedule length, cosine trajectory, and early-stopping patience, so the comparison is descriptive rather than a causal estimate of adding epochs. Individual networks may benefit from different learning rates, warmup, augmentation, regularization, distillation, input resolution, or longer schedules. The MobileNet scratch comparison measures behavior under one fixed budget and does not claim fully optimized scratch performance.

The pretrained benchmark remains confounded by upstream checkpoint provenance. RepViT uses a distilled checkpoint with dual classifier heads, and MobileNetV4-Conv-S uses a public timm checkpoint rather than an official Google checkpoint. The main comparison is therefore best understood as public model-package selection under common downstream conditions rather than pure causal isolation of architecture.

The datasets are all natural-image classification benchmarks, and Tiny ImageNet is related to the ImageNet pretraining source. The fixed 224 pixel adaptation does not test native-resolution behavior. The model pool is representative but not exhaustive and omits several recent lightweight CNNs. The resource analysis does not include quantization, energy, mobile-device execution, peak training memory, or optimized runtimes such as TensorRT, ONNX Runtime, and Core ML. The CPU benchmark includes three timing rounds, whereas the L4 distribution is based on one warmed sequence of 500 timed passes; the two environments therefore provide different forms of repeatability evidence.

\section{Conclusion}
Within the evaluated model pool, datasets, and execution environments, the results do not support a consistent advantage from architectural recency. EfficientNetV2-S records the highest observed CIFAR accuracy, RepViT-M1.0 leads Tiny ImageNet, EfficientNet-B0 belongs to every independently calculated bivariate accuracy-resource frontier, and MobileNetV3-Small provides the highest observed accuracy under the evaluated severe arithmetic and CPU-latency budgets.

The results also show why a single resource proxy is insufficient. GMACs correspond strongly to latency on the tested CPU but weakly to L4 latency. Peak CUDA tensor memory largely follows parameter count, while incremental allocation still reveals differences in activation and temporary-tensor demand. The fresh 100 epoch EfficientNet-B0 schedules improve over the separate shorter scratch schedules but leave substantial pretrained gaps, and the evaluated MobileNetV4-Conv-S package does not improve upon MobileNetV3-Small under the tested checkpoints or scratch protocol.

The appropriate conclusion is conditional rather than universal. Newer designs can improve particular datasets and deployment profiles, but model selection continues to depend on checkpoint provenance, target data, optimization budget, resource definition, and the complete execution environment.

\appendix
\section{Supplementary Predictive and Resource Results}
\label{app:supplementary}

\subsection{Pretrained top-5 accuracy}
\label{app:top5}
\begin{table}[H]
\centering
\caption{Pretrained top-5 accuracy. CIFAR-10 values are near saturation and are included for completeness.}
\label{tab:top5}
\scriptsize
\begin{tabular}{lrrr}
\toprule
\textbf{Model} & \textbf{CIFAR-10 (\%)} & \textbf{CIFAR-100 (\%)} & \textbf{Tiny ImageNet (\%)} \\
\midrule
EfficientNetV2-S & 99.84 & 96.69 & 91.79 \\
EfficientNet-B0 & 99.87 & \textbf{97.29} & 92.42 \\
RepViT-M1.0 & 99.82 & 96.58 & \textbf{93.09} \\
ResNet18 & 99.72 & 95.42 & 86.33 \\
MobileNetV3-Small & 99.76 & 96.48 & 89.32 \\
MobileNetV2 & 99.71 & 95.92 & 89.57 \\
MobileNetV4-Conv-S & 99.68 & 95.54 & 86.98 \\
ShuffleNetV2 x1.0 & 99.70 & 96.19 & 88.69 \\
SqueezeNet1.1 & 99.16 & 93.65 & 83.13 \\
\bottomrule
\end{tabular}
\end{table}

\subsection{Accuracy versus peak CUDA allocation}
\label{app:memoryplot}
\begin{figure}[H]
\centering
\includegraphics[width=\textwidth]{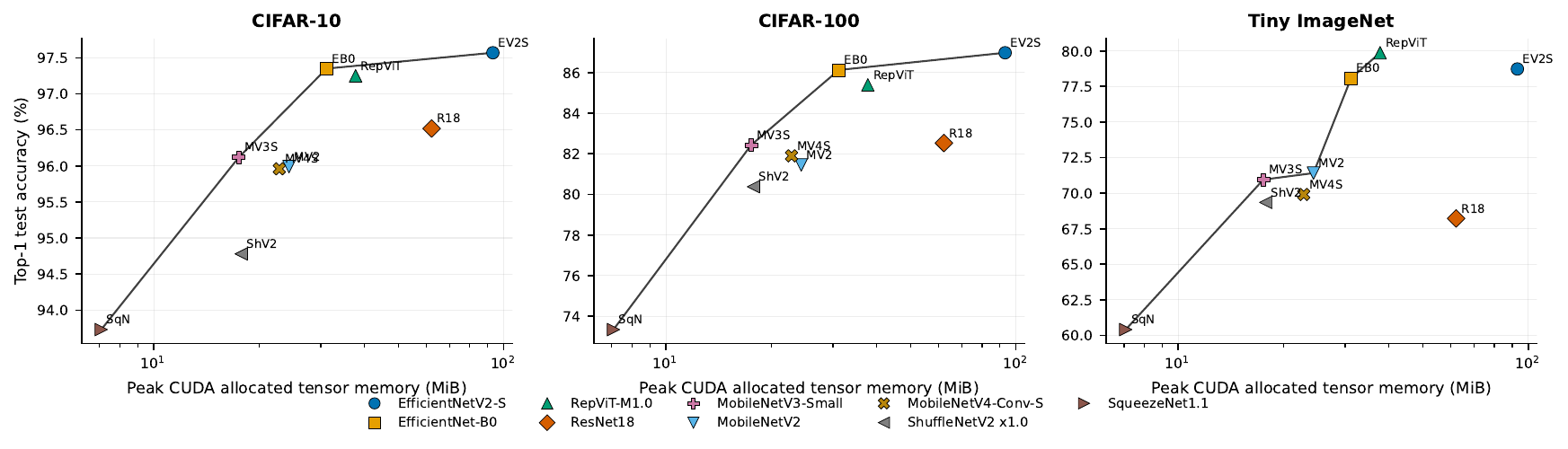}
\caption{Top-1 accuracy against peak PyTorch CUDA allocated tensor memory during batch size 1 FP16 autocast inference. Frontiers are calculated independently for each dataset.}
\label{fig:peakmemory}
\end{figure}

\subsection{Descriptive rank associations}
\label{app:correlations}
\begin{table}[H]
\centering
\caption{Descriptive Spearman rank associations across the nine evaluated models.}
\label{tab:correlations}
\scriptsize
\begin{tabular}{lr}
\toprule
\textbf{Comparison} & \textbf{Spearman $\rho$} \\
\midrule
GMACs versus L4 median latency & 0.27 \\
GMACs versus CPU one-thread median latency & 0.95 \\
GMACs versus CPU four-thread median latency & 0.93 \\
L4 versus CPU one-thread latency & 0.40 \\
L4 versus CPU four-thread latency & 0.50 \\
CPU one-thread versus CPU four-thread latency & 0.95 \\
Parameters versus peak CUDA allocated memory & 0.97 \\
CIFAR-10 versus CIFAR-100 accuracy rank & 0.98 \\
CIFAR-10 versus Tiny ImageNet accuracy rank & 0.77 \\
CIFAR-100 versus Tiny ImageNet accuracy rank & 0.73 \\
\bottomrule
\end{tabular}
\end{table}

\subsection{Selected paired comparisons}
\label{app:pairedtable}

\begin{table}[H]
\centering
\caption{Selected paired comparisons. Differences are model A minus model B in percentage points.}
\label{tab:paired}
\scriptsize
\resizebox{\textwidth}{!}{%
\begin{tabular}{llllrrr}
\toprule
\textbf{Setting} & \textbf{Dataset} & \textbf{Model A} & \textbf{Model B} & \textbf{Difference} & \textbf{95\% interval} & \textbf{McNemar $p$} \\
\midrule
Pretrained & CIFAR-10 & EfficientNetV2-S & EfficientNet-B0 & 0.22 & [-0.10, 0.54] & 0.196 \\
Pretrained & CIFAR-100 & EfficientNetV2-S & EfficientNet-B0 & 0.85 & [0.22, 1.48] & 0.0085 \\
Pretrained & Tiny ImageNet & RepViT-M1.0 & EfficientNetV2-S & 1.14 & [0.42, 1.85] & 0.0018 \\
Pretrained & Tiny ImageNet & RepViT-M1.0 & EfficientNet-B0 & 1.79 & [1.06, 2.51] & $<0.00001$ \\
Pretrained & CIFAR-10 & MobileNetV3-Small & MobileNetV4-Conv-S & 0.16 & [-0.24, 0.56] & 0.458 \\
Pretrained & CIFAR-100 & MobileNetV3-Small & MobileNetV4-Conv-S & 0.53 & [-0.16, 1.22] & 0.139 \\
Pretrained & Tiny ImageNet & MobileNetV3-Small & MobileNetV4-Conv-S & 1.04 & [0.23, 1.84] & 0.0124 \\
Scratch & CIFAR-10 & MobileNetV3-Small & MobileNetV4-Conv-S & 2.55 & [1.90, 3.21] & $<0.00001$ \\
Scratch & CIFAR-100 & MobileNetV3-Small & MobileNetV4-Conv-S & 1.76 & [0.86, 2.67] & 0.00015 \\
Scratch & Tiny ImageNet & MobileNetV3-Small & MobileNetV4-Conv-S & 0.99 & [0.14, 1.84] & 0.0255 \\
\bottomrule
\end{tabular}}
\end{table}

\section{Point Estimate Pareto Membership}
Table~\ref{tab:paretoappendix} summarizes the point estimate frontier members used in the main analysis.

\begin{table}[H]
\centering
\caption{Point estimate Pareto frontier members by dataset and resource metric.}
\label{tab:paretoappendix}
\scriptsize
\begin{tabularx}{\textwidth}{p{2.1cm}p{2.5cm}X}
\toprule
\textbf{Dataset} & \textbf{Resource} & \textbf{Frontier members in increasing resource order} \\
\midrule
CIFAR-10 & Parameters & SqueezeNet1.1, ShuffleNetV2 x1.0, MobileNetV3-Small, EfficientNet-B0, EfficientNetV2-S \\
CIFAR-10 & GMACs & MobileNetV3-Small, EfficientNet-B0, EfficientNetV2-S \\
CIFAR-10 & L4 median latency & ResNet18, EfficientNet-B0, EfficientNetV2-S \\
CIFAR-10 & CPU median latency & MobileNetV3-Small, EfficientNet-B0, EfficientNetV2-S \\
CIFAR-10 & Peak CUDA memory & SqueezeNet1.1, MobileNetV3-Small, EfficientNet-B0, EfficientNetV2-S \\
\midrule
CIFAR-100 & Parameters & SqueezeNet1.1, ShuffleNetV2 x1.0, MobileNetV3-Small, EfficientNet-B0, EfficientNetV2-S \\
CIFAR-100 & GMACs & MobileNetV3-Small, EfficientNet-B0, EfficientNetV2-S \\
CIFAR-100 & L4 median latency & ResNet18, EfficientNet-B0, EfficientNetV2-S \\
CIFAR-100 & CPU median latency & MobileNetV3-Small, EfficientNet-B0, EfficientNetV2-S \\
CIFAR-100 & Peak CUDA memory & SqueezeNet1.1, MobileNetV3-Small, EfficientNet-B0, EfficientNetV2-S \\
\midrule
Tiny ImageNet & Parameters & SqueezeNet1.1, ShuffleNetV2 x1.0, MobileNetV3-Small, MobileNetV2, EfficientNet-B0, RepViT-M1.0 \\
Tiny ImageNet & GMACs & MobileNetV3-Small, MobileNetV2, EfficientNet-B0, RepViT-M1.0 \\
Tiny ImageNet & L4 median latency & ResNet18, MobileNetV4-Conv-S, MobileNetV2, EfficientNet-B0, RepViT-M1.0 \\
Tiny ImageNet & CPU median latency & MobileNetV3-Small, MobileNetV2, EfficientNet-B0, RepViT-M1.0 \\
Tiny ImageNet & Peak CUDA memory & SqueezeNet1.1, MobileNetV3-Small, MobileNetV2, EfficientNet-B0, RepViT-M1.0 \\
\bottomrule
\end{tabularx}
\end{table}

\section{Additional Reproducibility Details}
Training augmentation uses bicubic random resized cropping with a scale of 0.80 to 1.00 and an aspect ratio of 0.90 to 1.10, horizontal flipping with probability 0.5, and color jitter values of 0.15 for brightness, contrast, and saturation and 0.05 for hue. Evaluation uses direct bicubic resizing to $224\times224$ without random cropping. ImageNet channel means and standard deviations are used for normalization. Training uses seeded data loading, \texttt{drop\_last=True}, deterministic cuDNN mode, and disabled cuDNN benchmarking. One multiply-accumulate operation is counted as one MAC and reported in GMACs accordingly.


\begin{thebibliography}{99}\small
\bibitem{lecun2015} Y. LeCun, Y. Bengio, and G. Hinton, ``Deep learning,'' \textit{Nature}, vol. 521, no. 7553, pp. 436--444, 2015.
\bibitem{imagenet} J. Deng, W. Dong, R. Socher, L. Li, K. Li, and L. Fei-Fei, ``ImageNet: A large-scale hierarchical image database,'' in \textit{Proc. IEEE Conf. Comput. Vis. Pattern Recognit.}, 2009, pp. 248--255.
\bibitem{cifar} A. Krizhevsky, ``Learning multiple layers of features from tiny images,'' University of Toronto, Technical Report, 2009.
\bibitem{tiny} Y. Le and X. Yang, ``Tiny ImageNet visual recognition challenge,'' Stanford CS231n Course Project, 2015.
\bibitem{resnet} K. He, X. Zhang, S. Ren, and J. Sun, ``Deep residual learning for image recognition,'' in \textit{Proc. IEEE Conf. Comput. Vis. Pattern Recognit.}, 2016, pp. 770--778.
\bibitem{squeezenet} F. N. Iandola et al., ``SqueezeNet: AlexNet-level accuracy with 50x fewer parameters and less than 0.5MB model size,'' arXiv:1602.07360, 2016.
\bibitem{mnv2} M. Sandler, A. Howard, M. Zhu, A. Zhmoginov, and L.-C. Chen, ``MobileNetV2: Inverted residuals and linear bottlenecks,'' in \textit{Proc. IEEE Conf. Comput. Vis. Pattern Recognit.}, 2018, pp. 4510--4520.
\bibitem{shuffle} N. Ma, X. Zhang, H.-T. Zheng, and J. Sun, ``ShuffleNet V2: Practical guidelines for efficient CNN architecture design,'' in \textit{Proc. Eur. Conf. Comput. Vis.}, 2018, pp. 116--131.
\bibitem{mnv3} A. Howard et al., ``Searching for MobileNetV3,'' in \textit{Proc. IEEE Int. Conf. Comput. Vis.}, 2019, pp. 1314--1324.
\bibitem{effnet} M. Tan and Q. Le, ``EfficientNet: Rethinking model scaling for convolutional neural networks,'' in \textit{Proc. Int. Conf. Mach. Learn.}, 2019, pp. 6105--6114.
\bibitem{effnetv2} M. Tan and Q. Le, ``EfficientNetV2: Smaller models and faster training,'' in \textit{Proc. Int. Conf. Mach. Learn.}, 2021, pp. 10096--10106.
\bibitem{mobileone} P. K. A. Vasu, J. Gabriel, J. Zhu, O. Tuzel, and A. Ranjan, ``MobileOne: An improved one millisecond mobile backbone,'' in \textit{Proc. IEEE/CVF Conf. Comput. Vis. Pattern Recognit.}, 2023, pp. 7907--7917.
\bibitem{fasternet} J. Chen et al., ``Run, Don't Walk: Chasing higher FLOPS for faster neural networks,'' in \textit{Proc. IEEE/CVF Conf. Comput. Vis. Pattern Recognit.}, 2023, pp. 12021--12031.
\bibitem{fastvit} P. K. A. Vasu, J. Gabriel, J. Zhu, O. Tuzel, and A. Ranjan, ``FastViT: A fast hybrid vision transformer using structural reparameterization,'' in \textit{Proc. IEEE/CVF Int. Conf. Comput. Vis.}, 2023, pp. 5785--5795.
\bibitem{repvit} A. Wang, H. Chen, Z. Lin, J. Han, and G. Ding, ``RepViT: Revisiting mobile CNN from ViT perspective,'' in \textit{Proc. IEEE/CVF Conf. Comput. Vis. Pattern Recognit.}, 2024, pp. 15909--15920.
\bibitem{mnv4} D. Qin et al., ``MobileNetV4: Universal models for the mobile ecosystem,'' in \textit{Proc. Eur. Conf. Comput. Vis.}, 2024, pp. 78--96.
\bibitem{starnet} X. Ma, X. Dai, Y. Bai, Y. Wang, and Y. Fu, ``Rewrite the Stars,'' in \textit{Proc. IEEE/CVF Conf. Comput. Vis. Pattern Recognit.}, 2024, pp. 5694--5703.
\bibitem{lsnet} A. Wang, H. Chen, Z. Lin, J. Han, and G. Ding, ``LSNet: See Large, Focus Small,'' in \textit{Proc. IEEE/CVF Conf. Comput. Vis. Pattern Recognit.}, 2025, pp. 9718--9729.
\bibitem{uniconvnet} Y. Wang and W. Xi, ``UniConvNet: Expanding effective receptive field while maintaining asymptotically Gaussian distribution for ConvNets of any scale,'' in \textit{Proc. IEEE/CVF Int. Conf. Comput. Vis.}, 2025, pp. 20922--20933.
\bibitem{kornblith} S. Kornblith, M. Norouzi, H. Lee, and G. Hinton, ``Do better ImageNet models transfer better?'' in \textit{Proc. IEEE Conf. Comput. Vis. Pattern Recognit.}, 2019, pp. 2661--2671.
\bibitem{rethinking} K. He, R. Girshick, and P. Doll\'ar, ``Rethinking ImageNet pre-training,'' in \textit{Proc. IEEE Int. Conf. Comput. Vis.}, 2019, pp. 4918--4927.
\bibitem{goldblum} M. Goldblum et al., ``Battle of the backbones: A large-scale comparison of pretrained models across computer vision tasks,'' in \textit{Advances in Neural Information Processing Systems}, vol. 36, Datasets and Benchmarks Track, 2023.
\bibitem{pegeot} T. P\'egeot, I. Kucher, A. Popescu, and B. Delezoide, ``A comprehensive study of transfer learning under constraints,'' in \textit{Proc. IEEE/CVF Int. Conf. Comput. Vis. Workshops}, 2023, pp. 1148--1157.
\bibitem{tiwari} R. Tiwari et al., ``RCV2023 challenges: Benchmarking model training and inference for resource-constrained deep learning,'' in \textit{Proc. IEEE/CVF Int. Conf. Comput. Vis. Workshops}, 2023, pp. 1534--1543.
\bibitem{jeevan} P. Jeevan and A. Sethi, ``Which backbone to use: A resource-efficient domain specific comparison for computer vision,'' \textit{Transactions on Machine Learning Research}, 2025.
\bibitem{guerin} J. Guerin, S. Bansal, A. Shaban, P. Mann, and H. Gazula, ``Vision backbone efficient selection for image classification in low-data regimes,'' in \textit{Proc. 36th British Mach. Vis. Conf.}, 2025, Paper 788.
\bibitem{shahriar2025} T. Shahriar, ``Comparative Analysis of Lightweight CNNs for Resource-Constrained Devices: Predictive Performance, Efficiency Trade-offs, and Initialization Effects,'' arXiv:2505.03303, 2025.
\bibitem{adamw} I. Loshchilov and F. Hutter, ``Decoupled weight decay regularization,'' in \textit{Proc. Int. Conf. Learn. Represent.}, 2019.
\bibitem{sgdr} I. Loshchilov and F. Hutter, ``SGDR: Stochastic gradient descent with warm restarts,'' in \textit{Proc. Int. Conf. Learn. Represent.}, 2017.
\bibitem{labelsmooth} C. Szegedy et al., ``Rethinking the Inception architecture for computer vision,'' in \textit{Proc. IEEE Conf. Comput. Vis. Pattern Recognit.}, 2016, pp. 2818--2826.
\bibitem{miettinen} K. Miettinen, \textit{Nonlinear Multiobjective Optimization}. Boston, MA, USA: Kluwer Academic Publishers, 1999.
\bibitem{spearman} C. Spearman, ``The proof and measurement of association between two things,'' \textit{The American Journal of Psychology}, vol. 15, no. 1, pp. 72--101, 1904.
\bibitem{mcnemar} Q. McNemar, ``Note on the sampling error of the difference between correlated proportions or percentages,'' \textit{Psychometrika}, vol. 12, pp. 153--157, 1947.
\bibitem{efron} B. Efron and R. J. Tibshirani, \textit{An Introduction to the Bootstrap}. New York, NY, USA: Chapman and Hall, 1993.
\bibitem{timm} R. Wightman, ``PyTorch Image Models,'' GitHub repository and Zenodo software archive, 2019.
\end{thebibliography}
\end{document}